\documentclass[runningheads]{llncs}
\usepackage[T1]{fontenc}
\usepackage[ruled,vlined,linesnumbered]{algorithm2e}
\usepackage{amsmath}
\usepackage{amssymb}
\usepackage[T1]{fontenc}
\usepackage{hyperref}
\usepackage{graphicx}
\usepackage{flafter}
\usepackage{placeins}
\usepackage{flafter}
\begin{document}
\title{Explaining Unsupervised Disease Staging in Huntington’s Disease: Insights into Model Representations and Clusters}
%
%
\author{Lubna Mahmoud Abu Zohair\orcidID{0000-0002-7686-5364} \and Hind Zantout\orcidID{0000-0002-3804-0513}}
\authorrunning{L. Abu Zohair and H. Zantout}
%
\institute{Heriot-Watt University, Dubai, United Arab Emirates}
\maketitle              
\begin{abstract}
Huntington’s disease (HD) is a progressive neurodegenerative disorder that affects motor, cognitive, and behavioral functions, where accurate characterization of disease progression remains essential to improve patient outcome and quality of life. Unsupervised machine learning (ML) approaches have demonstrated the ability to uncover disease progression trajectories and meaningful latents stages from longitudinal data; however, their limited interpretability restricts clinical trust and translation.

We extend a previously proposed ML-based disease staging framework by applying an explainability analysis to the extracted feature representations and discovered disease stages. Applied to the Enroll-HD dataset, we first project the learned representations into a lower-dimensional space to intuitively assess whether the resulting clusters align with the progression of established clinical measures. We then use saliency maps to identify the clinical features that most strongly contribute to the learned embeddings over time. Finally, we train a surrogate classifier and apply SHAP to quantify feature importance for cluster assignments and to analyze which clinical variables drive transitions between disease stages.

The explainability analysis indicates that the learned embeddings capture clinically meaningful disease structure, aligning with established motor and functional severity scores and exhibiting progressive deterioration across clusters. Within this analysis, SHAP reveals a stratification of disease stages, ranging from early cognitive–motor impairment to severe functional dependency, consistent with known clinical progression patterns, while also highlighting intra-stage variability.

Leveraging established and widely accessible methods, this work enhances trust in unsupervised modelling by providing interpretability of latent disease progression patterns, improving transparency and clinical interpretability, supporting data-driven staging on comprehensive datasets, and facilitating clinical feedback for reliable implementation. These findings may further generalize to other neurodegenerative conditions

\end{abstract}
\section{Introduction}
Huntington’s disease (HD) is an inherited neurodegenerative disorder that causes a progressive decline of nerve cells in the brain, leading to deterioration in motor, cognitive, and behavioural functions \cite{Jakel2000,JMAI10176}. Understanding how the disease progresses over time is essential for early diagnosis, prognosis, patient stratification for effective interventions, and the development of targeted therapies  \cite{JMAI10176,AbuZohair2025}. One way to achieve this is by identifying different stages of the disease to support an accurate diagnosis. Clinical efforts toward this goal have evolved, beginning with the ability to identify HD gene carriers, followed by the characterization of clinical manifestations through expert evaluation using measures such as the Unified HD Rating Scale (UHDRS) diagnostic confidence level (DCL) \cite{Romana2020}. Subsequently, this framework was extended by quantifying the severity of the disease based on functional impairment, particularly through the Total Functional Capacity (TFC) score of the UHDRS \cite{Shoulson1979}. To date, these remain among the standardized and widely used approaches for HD clinical characterization and staging; however, they have notable limitations \cite{Paulsen2025HDcriteria}. These methods are either domain specific (either derived from motor or functional tasks), rely heavily on subjective clinical judgment, or depend on predefined thresholds clinical measurements derived by consensus \cite{Winder20188,Paulsen2025HDcriteria}. Such thresholds impose rigid stage boundaries that can overlook intra-stage variability and fail to capture the continuous and heterogeneous nature of disease progression, as is evident in the Shoulson and Fahn staging system and recently proposed framework Huntington Disease Integrated Staging System \cite{Tabrizi2022,Shoulson1979,Winder20188,Paulsen2025HDcriteria}.

Unsupervised learning methods have shown promise in uncovering latent disease stages without relying on predefined clinical cut-offs. A recent proof-of-concept work has leveraged unsupervised machine learning (ML) to extract patterns and perform clustering, enabling data-driven identification of HD disease stages across multidomain clinical features \cite{URL-STFN2026}. A graph-based representation learning framework (URL-STFN) \cite{URL-STFN-Parkinson} combined with a clustering pipeline has been proposed and it was successful in identifying meaningful disease stages, demonstrating minimal overlap in feature distributions across the derived stages \cite{URL-STFN2026}. While the resulting stages exhibit clinically relevant characteristics, the framework remains largely opaque in how clinical measurements at each patient visit are encoded and how latent temporal patterns reflecting disease progression align with established clinical scores such as motor and functional measures. Moreover, it does not explicitly identify which features drive cluster formation or transitions between disease stages, limiting interpretability and clinical trust. 

In this work, our objective is to complement the proof-of-concept URL-STFN-based staging framework \cite{URL-STFN2026} by not only providing intuitive visual inspection of the cluster structure and their alignment with established aggregated clinical measures, but more importantly by incorporating clinical feature attribution and model explainability to address the following questions:
\begin{itemize}
    \item  Which clinical features most strongly influence the learned embeddings over ages?
    \item  Does proximity in the embedding space reflect clinically meaningful similarity across patient visits and age groups?
    \item  Which clinical variables drive cluster formation and transitions between disease stages?
\end{itemize}
Enhanced transparency in both the learned representations and clustering algorithms improves the utility of the proposed ML-based framework, strengthens trustworthiness, and increases confidence in data-driven disease staging while supporting its reproducibility. Ultimately, this contributes to more reliable and interpretable unsupervised ML–based staging frameworks for clinical research and potential clinical application.
\subsection {Overview of the Machine learning-Based Disease Staging Framework}

The disease staging framework consists of two main components: a graph-based representation learning model designed to capture structural and longitudinal patterns in patient trajectories, and a clustering module with stability assessment to identify a stable number of disease stages \cite{URL-STFN2026}. Patient trajectories are modelled using an age-based dynamic graph, where each graph snapshot corresponds to a specific age, nodes represent patient visits, and edges encode similarity between clinical profiles using cosine similarity. To reduce noise and retain meaningful relationships, only strong similarities, defined by a threshold based on the 90th percentile of pairwise similarities, were preserved. This formulation enables the modelling of both intra-age clinical profiles relationships and their longitudinal progression. To learn these patterns, a graph-based representation learning framework, referred to as URL-STFN, was employed\cite{URL-STFN-Parkinson}. Specifically, graph convolutional networks were used to encode clinical feature patterns at each visit by capturing structural relationships among patients within the same age group. These spatial representations were then integrated with temporal dynamics across visits using a Gated Recurrent Unit, allowing the model to capture both within-visit feature interactions and longitudinal progression patterns. The resulting unified embeddings represent each clinical patient visit in a compact latent space that reflects disease progression dynamics without relying on predefined staging criteria. The model was trained in an unsupervised manner by reconstructing node features using a mean squared error loss, with grid search used to identify a stable and compact latent representation. 

The learned embeddings were clustered using K-means++, with the optimal number of clusters selected based on standard clustering metrics. The best-performing solution achieved strong separation with two clusters (Silhouette Score = 0.670 ± 0.027, Davies–Bouldin Index = 0.568 ± 0.037, Calinski–Harabasz Index = 453.98 ± 75.25). Stability analysis further indicated that the representation supports up to four robust clusters corresponding to potential disease stages, which were confirmed to be statistically distinct (PERMANOVA, p = 0.001).

The approach was applied to a curated subset of the Enroll-HD dataset, where only patients with at least four consecutive annual clinical visits at shared ages were included. This constraint ensured consistent longitudinal coverage across patients, reducing bias due to irregular or missing follow-ups. The resulting cohort comprised 302 patients with over one thousand longitudinal observations, spanning an age range of 54 to 57 years and a mean CAG repeat length of 42.5 ± 1.9. The dataset included both premanifest (19.4\%) and manifest (80.5\%) individuals. A total of 44 clinical features were selected across motor, cognitive, and functional domains, reflecting commonly used measurements in HD staging studies. For the implementation of the proof-of-concept, the missing values were explicitly encoded as $\neg 1$ to preserve the missing information, and all features were standardized to ensure comparability across heterogeneous clinical scales. The aggregated clinical scores (such as TFC, DCL, and motor scores) were excluded from the model input to maintain staging objectivity; however, they are used in this work to support result validation and enhance interpretability.

The clinical features showed a progressive and clinically consistent decline across the discovered stages, aligning with increasing HD severity and established disease characteristics, while also capturing additional intra-stage variability not reflected in traditional staging systems. These findings provide a data-driven and objective foundation for modelling HD progression and motivate the development of explainability extensions in this paper to advance this proof of concept into more comprehensive work, as a first step toward clinical translation of a ML–based, data-driven HD staging system.

\section{Method}

Despite the meaningful structure of the discovered stages, the mapping between clusters and established  clinical severity measurement has not been explicitly established, and it remains unclear how clinical variables are encoded or which features drive cluster formation and transitions between stages. Therefore, we first performed an exploratory analysis of the learned representations by projecting the learned clinical visits' embeddings into two dimensions using Uniform Manifold Approximation and Projection (UMAP) \cite{UMAP2020}. The embedding space was visualized in the scatter plot by colouring points according to cluster assignments to assess the separability of the disease stage and according to established clinical scores (motor and functional) to evaluate discovered stages alignment with clinical severity metrics. Subsequently, a saliency map is applied to attribute latent representations to input clinical features over time. This enables identification of the variables that most strongly influence the embedding and provides insight into how temporal clinical dynamics are encoded in the latent space. After that, SHapley Additive exPlanations (SHAP) is used to quantify feature contributions and allows the identification of the clinical variables that drive cluster membership, differentiate disease stages, and influence transitions between stages over time.

Data used in this work were generously provided by the participants in the Enroll-HD study and made available by the CHDI Foundation, Inc. Enroll-HD is a global clinical research platform designed to facilitate clinical research in HD. Core datasets are collected annually from all research participants as part of this multi-center longitudinal observational study. Data are monitored for quality and accuracy using a risk-based monitoring approach. All sites are required to obtain and maintain local ethical approval. In particular, patients’ longitudinal records in the Enroll table in Enroll-HD PDS7-R1 clinical dataset were mainly utilized in providing the concept of the proposed model. This table initially comprised around 30,511 unique participants and  129,537 recorded visits \cite{EnrollHD2025,EnrollHDPDS7Overview2025,CHDIFoundation}.
\subsection{Explaining Latent Representation}
Using gradient-based saliency maps, a common technique in computer vision and deep learning for model interpretability \cite{SaliencyMaps2014}, we quantify the importance of clinical features over time. The goal is to measure how each input features at each age influences the learned representations produced by the model.
For each patient age $t$, we copy the full input sequence and set features at time $t$; $x_t$, to require gradients, ensuring that gradients are computed only with respect to features at that age (steps 3-5; Algorithm~\ref{alg:saliency}).

The modified sequence is then passed through the model to obtain the fused embeddings (latent; step 6 in Algorithm~\ref{alg:saliency}). Since the model does not output a single scalar value, we define a differentiable proxy objective as the average $L_2$-norm $(\| h_t^{(i)} \|_2) $ of the embeddings at age $t$, which serves as a scalar measure of the embedding magnitude ($\mathcal{L}_t$). Back-propagation is then performed on this target to compute gradients with respect to the input features at age $t$. The absolute values of these gradients are taken as feature importance scores ($S_t$). These scores are averaged across all nodes (that correspond to patients) to obtain a single feature-level importance vector for each age ($\bar{S}_t$) (steps 8 and 9; Algorithm~\ref{alg:saliency}). Repeating this procedure for all ages yields a saliency  matrix $S$ of shape $T$×$F$, where each entry reflects how strongly a given clinical feature at a specific point in time contributes to the model’s representation. The matrix was transposed and normalized for heatmap visualization.
\begin{algorithm}[!b]
\caption{Saliency Estimation}
\label{alg:saliency}
\KwIn{$x \in \mathbb{R}^{T \times N \times F}$, model $f(\cdot)$, edge indices $\{\text{edge}_t\}$}
\KwOut{$S \in \mathbb{R}^{T \times F}$}
Initialize $\mathcal{S} \leftarrow [\,]$\

\For {$t = 1$ \KwTo $T$}{
    $x^{(t)} \leftarrow \text{copy and detach }(x)$\
    
    enable and compute gradients for $x_t$
    
    $x^{(t)}[t] \leftarrow x_t$

    $h_t \leftarrow f(x^{(t)})$

    $\mathcal{L}_t \leftarrow \frac{1}{N} \sum_{i=1}^{N} \| h_t^{(i)} \|_2$

    $S_t \leftarrow \left| \nabla_{x_t} \mathcal{L}_t \right|$

    $\bar{S}_t \leftarrow \frac{1}{N} \sum_{i=1}^{N} S_t^{(i)}$

    Append $\bar{S}_t$ to $\mathcal{S}$
}
\Return $\mathcal{S}$
\end{algorithm}
\subsection{Interpretation of Latent Clustering Structure}
Since clustering has no gradients and thus does not support direct feature attribution, we trained a surrogate Random Forest classifier to estimate the probability of the cluster assignment $y$ for URL-STFN embeddings $h$ (step 1; Algorithm~\ref{alg:shap_clustering}) \cite{clustsurrogate2024}. The wrapper function $g$ utilized to implement a two-stage mapping from input features to cluster probabilities, where input features are first transformed into latent representations by the model, and these representations are then mapped to cluster probabilities by the trained classifier. This formulation allows SHAP to estimate how variations in input features influence cluster assignments through the learned latent representations.
To estimate feature contributions over time, SHAP values were computed independently at each age to capture temporal dynamics. SHAP quantifies how much each feature shifts the model output away from a “typical patient” prediction, defined with respect to a background reference distribution. Specifically, 100 randomly sampled patients $B_t$ were used at each age as the background set, enabling SHAP to approximate feature contributions relative to this baseline. For a given age t, the corresponding node feature matrix $X_t$ was extracted, where each row represents an instance and each column corresponds to a feature. This explains steps 3-6 in Algorithm~\ref{alg:shap_clustering}.
Feature contributions were aggregated by computing the mean absolute SHAP values in patients (steps 7-11; Algorithm~\ref{alg:shap_clustering}) . These matrices were normalized using min–max scaling, transposed, and visualized using a heatmap, with row clustering enabled to identify groups of features with similar temporal contribution patterns.

To analyze how clinical feature importance changes across disease progression stages, we computed SHAP-based transition effects ($\Delta$ SHAP). After obtaining cluster-specific SHAP values, we quantified the change in feature contributions between consecutive disease stages by computing pairwise differences in SHAP importance scores across clusters.
Formally, for each clinical feature \( x \), we define Equation~\ref{eq:delta_shap}:
\begin{equation}
\Delta \text{SHAP}_{c_i \rightarrow c_j}(x)
= \text{SHAP}_{c_j}(x) - \text{SHAP}_{c_i}(x)
\label{eq:delta_shap}
\end{equation}
This results in a transition matrix capturing how feature importance evolves across stages (i.e. 0→1, 1→2, 2→3). Features were then ranked based on the maximum absolute change across all transitions to identify the strongest drivers of disease progression. The resulting $\Delta$SHAP matrix was visualized using a diverging heatmap to highlight features that increase or decrease in importance across disease stages, enabling interpretation of temporal shifts in clinical determinants of disease progression.

\section {Results}
\begin{figure}[!t]
\includegraphics[width=\textwidth]{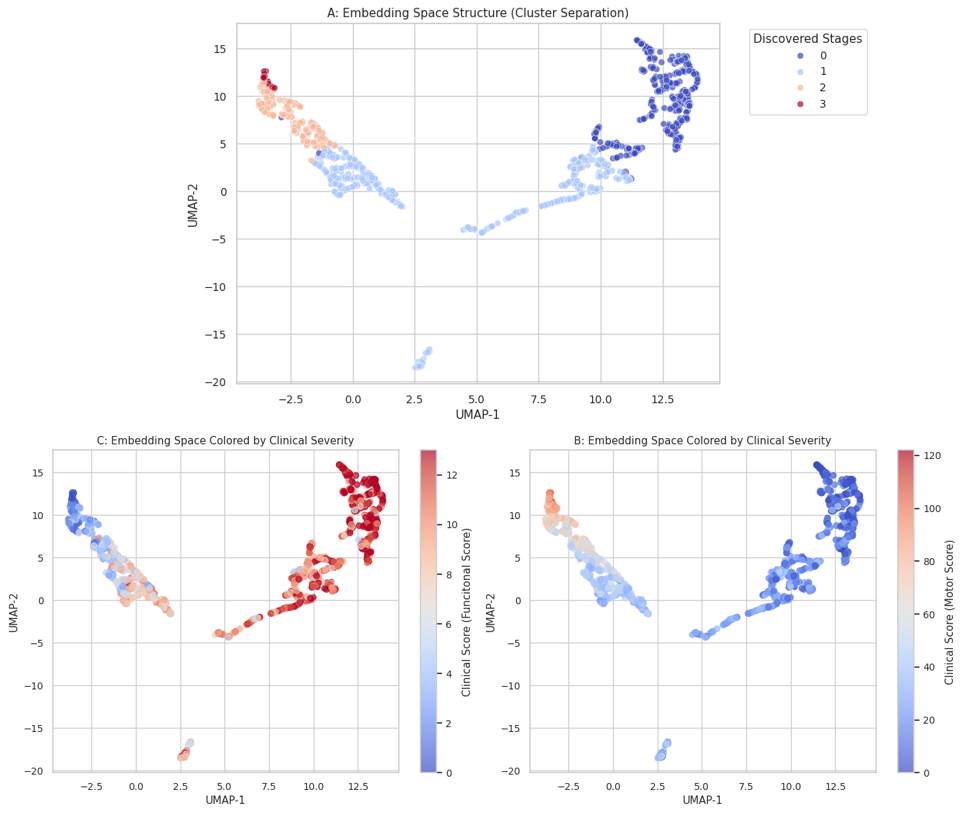}
\caption{
UMAP projection of the learned embeddings, coloured by (A) the corresponding cluster number, and clinical severity scores for (B) motor and (C) functional measures. 
}
\label{fig1}
\end{figure}

The distribution of clinical severity scores follows a pattern consistent with the progression of severity of the clusters discovered, with both motor and functional measures showing gradual deterioration throughout the embedding space, as illustrated in Fig.~\ref{fig1}. The cluster labels (0–3) do not reflect an intrinsic ordering from the clustering algorithm, as they are assigned arbitrarily in an unsupervised setting. For interpretability, clusters were ordered post hoc based on clinical characteristics and transition patterns observed in the data.This alignment suggests that the learned representations capture meaningful disease severity dynamics, as reflected by established clinical scales.

\begin{algorithm}[!h]
\caption{SHAP-Based Interpretation of Latent Clustering}
\label{alg:shap_clustering}
\KwIn{Input data $\{X_t\}_{t=1}^{T}$, model $f(\cdot)$, cluster labels $y$}
\KwOut{SHAP importance matrix $S \in \mathbb{R}^{F \times T}$}
Train Random Forest classifier $g$ on embeddings $h$ and labels $y$\;
\For{$t = 1$ \KwTo $T$}{
    Extract features $X_t$\;
    Sample background $B_t \subset X_t$\;
    Define wrapper:
    $$x \mapsto g(f(x))$$
    Compute SHAP values:
    $$\phi_t = \text{SHAP}(X_t; B_t)$$
}
\For{each cluster $c = 1$ \KwTo $K$}{
    \For{$t = 1$ \KwTo $T$}{
        Extract $\phi_t^{(c)} \in \mathbb{R}^{N \times F}$\;
        Compute mean importance:
       $$ S_t^{(c)} = \frac{1}{N} \sum_{i=1}^{N} |\phi_{t,i}^{(c)}|$$
    }
    Stack over time:
    $S^{(c)} \in \mathbb{R}^{F \times T}$\;
}
\Return $S$\;
\end{algorithm}

\subsection{Feature Attribution in Latent Representations}
\begin{figure}[!t]
\includegraphics[width=\textwidth]{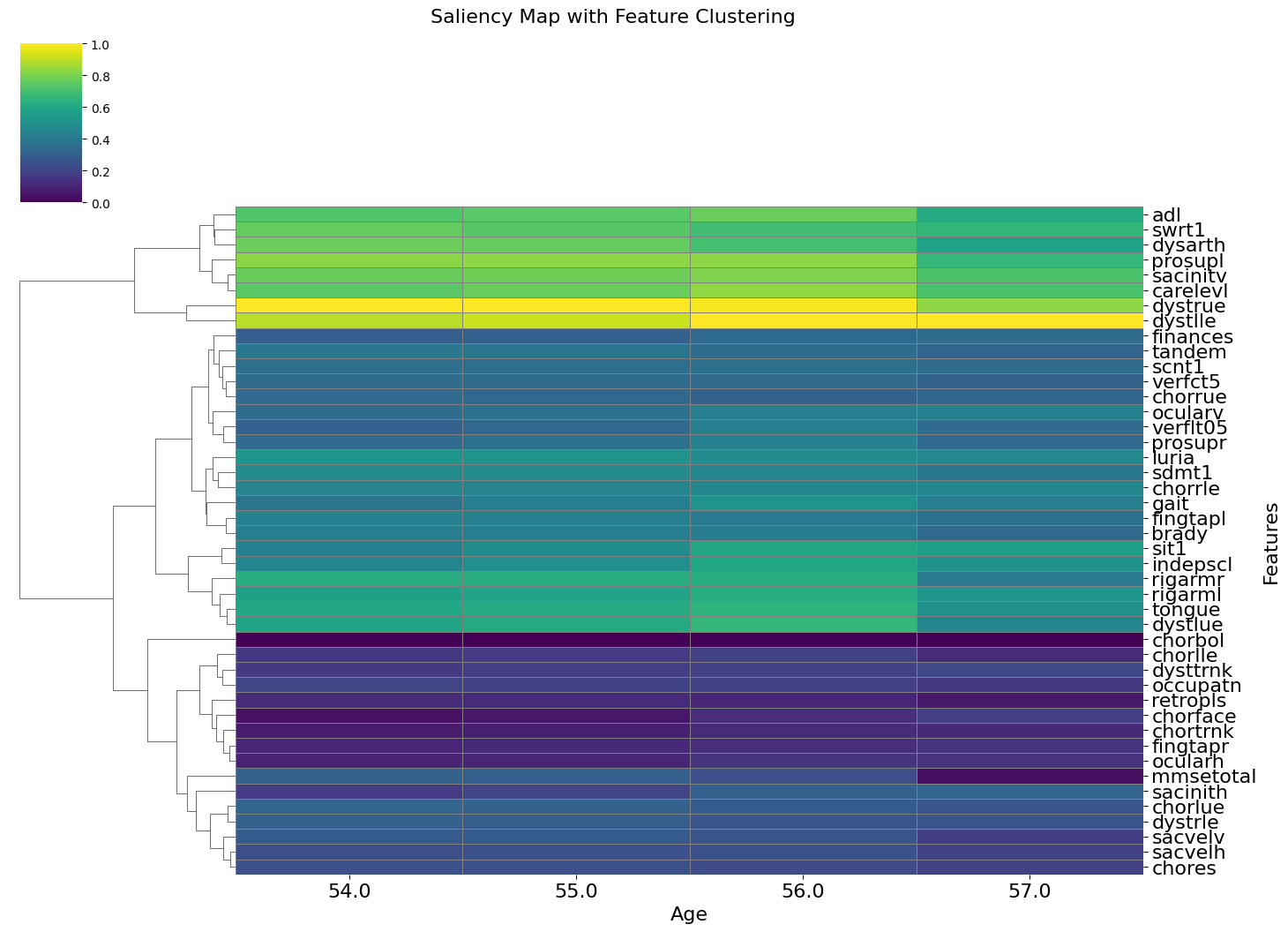}
\caption{
Saliency map for clinical features importance across age. Rows represent input features, while columns represent age/time points. Feature importance values are normalized, and hierarchical clustering is applied to group features with similar temporal saliency patterns.
}
\label{fig2}
\end{figure}

The heatmap presented in Fig.~\ref{fig2} provides a saliency-based interpretation of the importance of clinical measurements across ages within the URL-STFN-derived embedding space. Each cell represents the average contribution of a given clinical feature (averaged over all patients) to the learned representation at a specific age. The accompanying hierarchical clustering groups features with similar temporal importance patterns, revealing structured relationships within the embedding space.
Overall, the model assigns consistently higher importance to functional and motor-related features such as adl (activities of daily living, a functional measure), swrt1 (stroop word reading, a cognitive measure), and motor measures (like dysarthria and dystonia measures), indicating that these variables most strongly influence the learned embeddings over time. In contrast, several cognitive and task-based measures such as SDMT1, Stroop reading tasks, and verbal fluency features show lower and more variable importance over time, suggesting a more context-dependent contribution to the embedding. 

\subsection{Feature Attribution Across Clusters and Transitions}

\begin{figure}[!h]
\includegraphics[width=\textwidth]{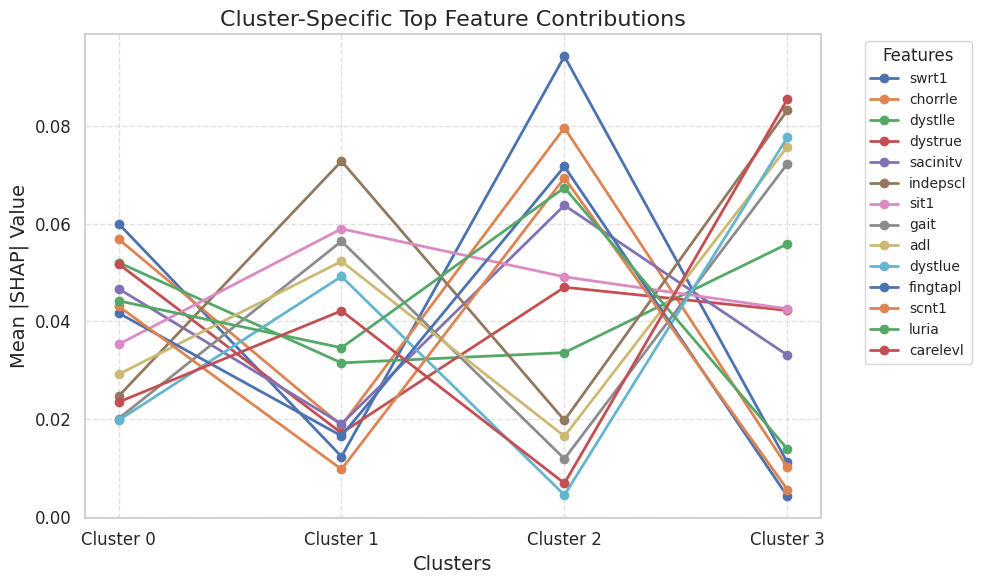}
\caption{
Cluster-specific clinical measurement importance based on SHAP analysis. For each cluster, feature contributions were quantified by averaging the absolute SHAP values of samples assigned to that cluster, using the SHAP values corresponding to the predicted cluster class. The top features, selected based on their overall importance, are shown. Each line represents a feature, and values indicate its relative contribution to cluster assignment. SHAP values were aggregated across samples (and ages) and normalized for reporting.
}
\label{fig3}
\end{figure}

Across all clusters, SHAP analysis revealed a clear stratification of disease stages along a continuum from mild cognitive and motor deficits (Cluster 0) to severe motor and functional impairment (Cluster 3), with intermediate phenotypes represented by Clusters 1 and 2, as illustrated in Fig.~\ref{fig3}. All reported SHAP importance values were normalized using min–max scaling in the range [0, 1] for comparability across features and clusters.

Cluster 0 was dominated by motor, coordination, and speech-related impairments, where the most influential feature was swrt1 (normalized SHAP importance = 1.00), followed by sacinitv (0.79), chorrle (0.78), dystrue (0.73), and dystlle (0.72). Additional contributions from dysarth (0.68) and scnt1 (0.67) indicate substantial involvement of speech and motor coordination deficits. Cognitive processing (sdmt1, 0.68) and fine motor control (fingtapl, 0.65) also contributed, suggesting a moderate disease stage with prominent motor–speech impairment. This is consistent with an early manifest disease stage characterized clinically by emerging chorea and speech/coordination deficits \cite{Ross2014}. In contrast, Cluster 1 was characterized by strong contributions from functional and mobility-related variables, indicating preserved independence. The most influential feature was indepscl (1.00), followed by sit1 (0.86), gait (0.78), and ocularv (0.69), reflecting a sustained motor function and daily activity performance. Additional contributions from adl (0.66) and carelevl (0.59) further support the interpretation of this cluster as representing a mild or early-stage phenotype with minimal functional decline. 

With more severe impairments, Cluster 2 exhibited a mixed phenotype characterized by concurrent cognitive, motor, and coordination deficits. While swrt1 remained the dominant feature (1.00), other highly ranked variables included chorrle (0.83), luria (0.72), fingtapl (0.71), and scnt1 (0.69). Speech-related impairment (dysarth, 0.62) and cognitive measures (sdmt1, 0.66) were also consistently influential. This pattern indicates multi-domain involvement, where cognitive and motor dysfunction co-occur, representing an intermediate stage between motor-dominant and functionally severe phenotypes, and consistent with known concurrent deterioration in motor control and coordination during disease progression and is also supported clinically \cite{Huntington2015}. In contrast, Cluster 3 was characterized by substantial functional dependency and advanced disease severity. The most influential feature was carelevl (0.96), followed by indepscl (0.91) and adl (0.77), indicating marked loss of independence. Motor impairment features such as dystlue (0.76), gait (0.73), and dystlle (0.71) were also highly ranked, alongside bulbar involvement reflected by tongue (0.57). This combination of features is consistent with a late-stage clinical phenotype characterized by severe functional and motor impairment \cite{Ross2014,Huntington2015}. Notably, the magnitude of SHAP values for these top contributing features also increases progressively with time, supporting the notion that the identified clusters capture a biologically and clinically meaningful progression, where feature importance not only differs across stages but also intensifies in line with advanced disease stages.
\begin{figure}[!b]
\includegraphics[width=0.99\textwidth]{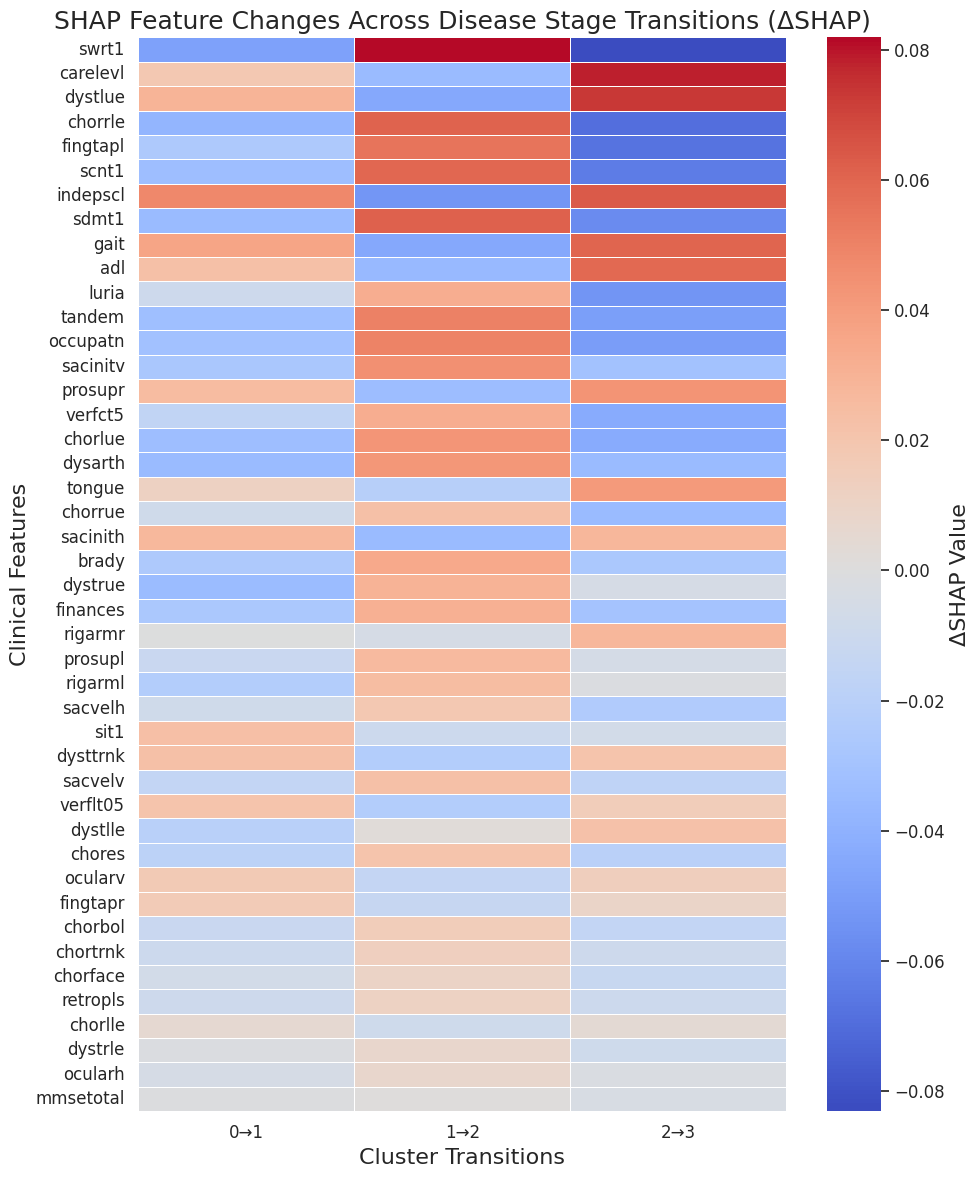}
\caption{
SHAP-based feature transition analysis across disease stages.
Heatmap showing $\Delta$SHAP values (difference in mean absolute SHAP importance) for all clinical features across consecutive cluster transitions (0→1, 1→2, 2→3). Positive values (red) indicate increasing feature influence, while negative values (blue) indicate decreasing influence across stages.
}
\label{fig4}
\end{figure}

Fig.~\ref{fig4} further reveals the drivers of disease transitions between clusters. For instance, the transition from Cluster 0 to Cluster 1 is primarily associated with increased importance of functional and motor decline measurements, including indepscl and gait, alongside early changes in dystonia-related features. Building  on this initial stage, the transition from Cluster 1 to Cluster 2 is further driven by motor impairment but shifts toward the inclusion of cognitive deterioration, with features such as swrt1, chorrle, scnt1, and sdmt1 showing the largest positive changes, indicating their growing influence in defining more advanced disease stages. Finally, the transition from Cluster 2 to Cluster 3 is dominated by functional dependency and overall disease burden indicators, particularly carelevl, indepscl, and adl, reflecting progression toward severe disability and loss of independence.

Overall, the feature attribution analyses reveal that cluster formation and disease transitions are driven by two dominant and related dimensions of disease severity. The first dimension is characterized by motor impairments, including swallowing difficulties, speech dysfunction, choking-related symptoms, and fine motor deficits (swrt1, chorrle, scnt1, fingtapl), which primarily distinguish Clusters 0 and 2. The second dimension is characterized by functional dependence and disability, including loss of independence, gait impairment, reduced activities of daily living, caregiver support requirements, and respiratory dysfunction (indepscl, gait, adl, carelevl, dystlue, dystrue), which primarily distinguish Clusters 1 and 3. Transition analysis further indicates that progression is associated with a shift in the relative importance of these feature groups. In particular, increases in motor and bulbar impairments are most prominent during the transition to Cluster 2, whereas progression to Cluster 3 is driven predominantly by increasing functional dependence and respiratory burden. These findings suggest that disease progression is not simply the accumulation of symptoms but rather reflects a transition from motor-dominant impairment toward broader functional disability, with motor and bulbar deficits emerging before becoming translated into substantial loss of independence and advanced disease burden.

\section{conclusion}
The explainability analysis of the ML-based disease staging framework \cite{URL-STFN2026} demonstrates that the learned representations and resulting clusters capture meaningful clinical structure, reflecting disease progression and showing alignment with established clinical scores, thereby supporting the clinical relevance of the model. Feature attribution further reveals the factors driving cluster formation and transitions, which are consistent with known patterns of disease severity progression. However, the present study focused on patients aged 54–57 years with at least four consecutive visits, which may limit the generalizability of the findings to broader HD populations. In addition, although missing values were encoded as $\neg$1 to preserve missingness information and avoid excluding incomplete records, future work should systematically compare this approach with alternative imputation and missing-data handling strategies. Further validation on larger and more diverse cohorts, will be essential to improve generalizability, strengthen clinical interpretation, and support the development of robust and objective ML-based staging frameworks for HD.

\begin{credits}
\subsubsection{\ackname} Data used in this work was generously provided by the participants in the Enroll-HD study and made available by CHDI Foundation, Inc. Enroll-HD is a global clinical research platform intended to accelerate progress towards therapeutics for Huntington's disease; core datasets are collected annually on all research participants as part of this multi-center longitudinal observational study. Enroll-HD is sponsored by CHDI Foundation, Inc., a nonprofit biomedical research organization exclusively dedicated to developing therapeutics for Huntington's disease. Enroll-HD would not be possible without the vital contribution of the research participants and their families \cite{CHDIFoundation}.
\subsubsection{\discintname}
The authors have no competing interests to declare that are
relevant to the content of this article.
\subsubsection{Data and Code Availability}
The Enroll-HD dataset used in the assessment of the proposed graph model can be requested directly from the Enroll-HD clinical research platform (\url{https://www.enroll-hd.org/}). The code and explainability analyses used to validate the proof-of-concept deployment of URL-STFN are publicly available at \url{https://github.com/LubnaM/URL-STFN_ExplainPoC/}. For further information regarding reproducibility and implementation details, contact the corresponding author.
\end{credits}
%
%
%
%

\end{document}